\documentclass[journal]{IEEEtran}
\usepackage{times}
\usepackage{epsfig}
\usepackage{graphicx} 
\usepackage{amsmath}
\usepackage{amssymb}
\usepackage{url}
\usepackage{caption}
\usepackage{cuted}
\usepackage{capt-of}
\usepackage{animate}
\usepackage{subfigure}

\usepackage{xcolor}
\usepackage[pagebackref=true,breaklinks=true,letterpaper=true,colorlinks,bookmarks=false]{hyperref}
\ifCLASSINFOpdf
\else
\fi
\begin{document}
%
\title{EnlightenGAN: Deep Light Enhancement without Paired Supervision}
%
%
%

\author{Yifan~Jiang,
        Xinyu~Gong,
        and~Ding~Liu,
        Yu~Cheng,
        Chen~Fang,
        Xiaohui~Shen,\\
        Jianchao~Yang,
        Pan~Zhou,
        and~Zhangyang Wang
\thanks{Y. Jiang, X. Gong and Z. Wang are with the Department of Electrical and Computer Engineering, the University of Texas at Austin, TX, USA. (email: yifanjiang97@utexas.edu; xinyu.gong@utexas.edu; atlaswang@utexas.edu)}
\thanks{D. Liu, C. Fang, X. Shen and J. Yang are with Bytedance Inc. (email: liuding@bytedance.com; fangchen@bytedance.com; shenxiaohui.kevin@bytedance.com;  yangjianchao@bytedance.com)}
\thanks{Y. Cheng is with Microsoft AI \& Research, Redmond, Washington 98052,
USA. (e-mail: yu.cheng@microsoft.com)}
\thanks{P. Zhou is with Department of Electronic Information and Communication, Huazhong University of Science and Technology. (email: panzhou@hust.edu.cn)}
\thanks{Correspondence addressed to: Zhangyang Wang (atlaswang@utexas.edu)}}

%
%

\markboth{Journal of \LaTeX\ Class Files,~Vol.~14, No.~8, August~2015}%
{Shell \MakeLowercase{\textit{et al.}}: Bare Demo of IEEEtran.cls for IEEE Journals}
%



\maketitle
\begin{figure*}[h!]
  \centering
  \includegraphics[width=17.5cm,height=6cm]{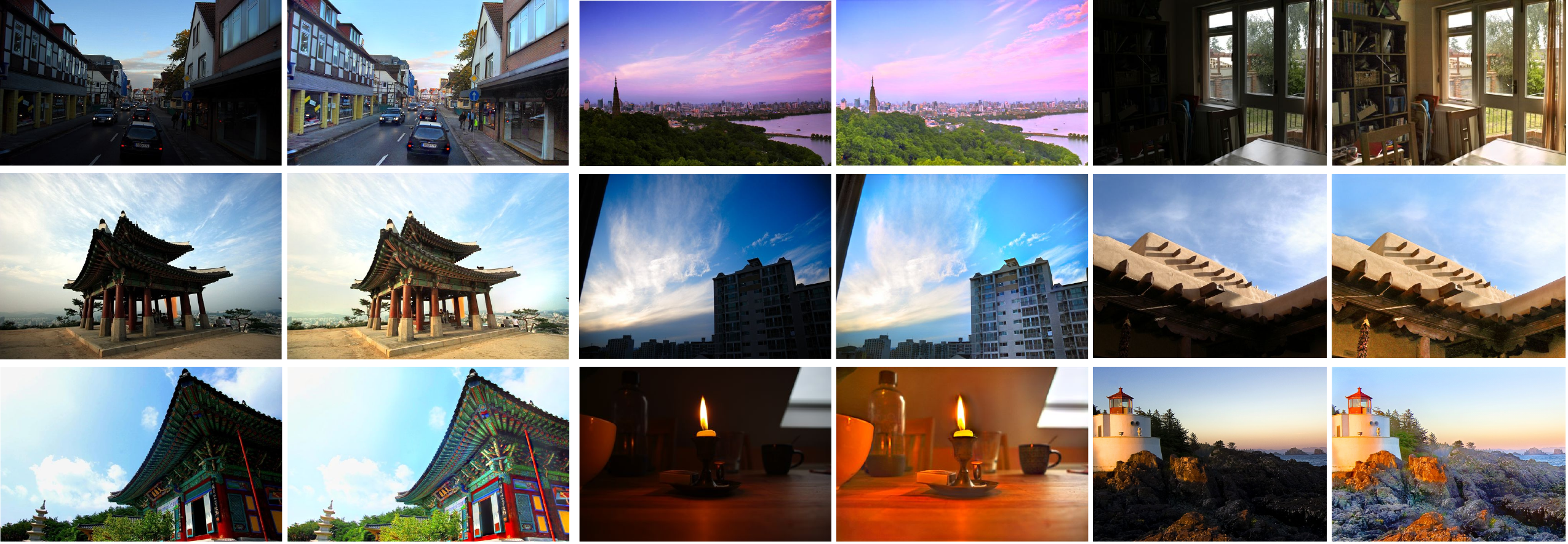}
  \caption{Representative visual examples by enhancing low-light images using EnlightenGAN. From left to right: columns 1, 3 and 5 are the  low-light input images; while columns 2, 4, an 6 their corresponding enhanced images by EnlightenGAN.}
\end{figure*}%

\begin{abstract}
Deep learning-based methods have achieved remarkable success in image restoration and enhancement, but are they still competitive when there is a lack of paired training data? As one such example, this paper explores the low-light image enhancement problem, where in practice it is extremely challenging to simultaneously take a low-light and a normal-light photo of the same visual scene. We propose a highly effective unsupervised generative adversarial network, dubbed \textit{EnlightenGAN}, that can be trained without low/normal-light image pairs, yet proves to generalize very well on various real-world test images. Instead of supervising the learning using ground truth data, we propose to regularize the unpaired training using the information extracted from the input itself, and benchmark a series of innovations for the low-light image enhancement problem, including a global-local discriminator structure, a self-regularized perceptual loss fusion, and the attention mechanism. Through extensive experiments, our proposed approach outperforms recent methods under a variety of metrics in terms of visual quality and subjective user study. Thanks to the great flexibility brought by unpaired training, EnlightenGAN is demonstrated to be easily adaptable to enhancing real-world images from various domains. Our codes and pre-trained models are available at: \url{https://github.com/VITA-Group/EnlightenGAN}.
\end{abstract}


%

\section{Introduction}
%
%
%
%

\IEEEPARstart{I}{Mage} captured in low-light conditions suffer from low contrast, poor visibility and high ISO noise. Those issues challenge both human visual perception that prefers high-visibility images, and numerous intelligent systems relying on computer vision algorithms such as all-day autonomous driving and biometric recognition \cite{yu2018bdd100k}. 
To mitigate the degradation, a large number of algorithms have been proposed, ranging from histogram or cognition-based ones \cite{land1977retinex,pizer1987adaptive} to learning-based approaches \cite{lore2017llnet,wei2018deep}. The state-of-the-art image restoration and enhancement approaches using deep learning heavily rely on either synthesized or captured corrupted and clean image pairs to train, such as super-resolution \cite{kim2016accurate}, denoising \cite{zhang2017beyond} and deblurring \cite{tao2018scale}. 

However, the availability assumption of paired training images has raised more difficulties, when it comes to enhancing images from more uncontrolled  scenarios, such as dehazing, deraining or low-light enhancement: 1) it is very difficult or even impractical to simultaneously capture corrupted and ground truth images of the same visual scene (e.g., low-light and normal-light image pairs at the same time); 2) synthesizing corrupted images from clean images could sometimes help, but such synthesized results are usually not photo-realistic enough, leading to various artifacts when the trained model is applied to real-world low-light images; 
3) specifically for the low-light enhancement problem, there may be no unique or well-defined high-light ground truth given a low-light image. For example, any photo taken from dawn to dusk could be viewed as a high-light version for the photo taken over the midnight at the same scene. 
Taking into account the above issues, our overarching goal is to enhance a low-light photo with spatially varying light conditions and over/under-exposure artifacts, while the paired training data is unavailable. 

Inspired by \cite{zhu2017unpaired,liu2017unsupervised} for unsupervised image-to-image translation, we adopt generative adversarial networks (GANs) to build an unpaired mapping between low and normal light image spaces without relying on exactly paired images. That frees us from training with only synthetic data or limited real paired data captured in controlled settings. We introduce a lightweight yet effective one-path GAN named \textbf{EnlightenGAN}, without using cycle-consistency as prior works  \cite{madam2018unsupervised, huang2018multimodal, choi2018stargan, hoffman2017cycada} and therefore enjoying the merit of much shorter training time. 

Due to the lack of paired training data, we 
incorporate a number of innovative techniques. 
We first propose a dual-discriminator to balance global and local low-light enhancement. Further, owing to the absence of ground-truth supervision, a self-regularized perceptual loss is proposed to constrain the feature distance between the low-light input image and its enhanced version, which is subsequently adopted both locally and globally together with the adversarial loss for training EnlightenGAN.
We also propose to exploit the illumination information of the low-light input as a self-regularized attentional map in each level of deep features to regularize the unsupervised learning.
Thanks to the unsupervised setting, we show that EnlightenGAN can be very easily adapted to enhancing real-world low-light images from different domains.

We highlight the notable innovations of EnlightenGAN: 

\begin{itemize}
 \vspace{0.5em}
  \item EnlightenGAN is the \textbf{first work} that successfully introduces unpaired training to low-light image enhancement. Such a training strategy removes the dependency on paired training data and enables us to train with larger varieties of images from different domains. It also avoids overfitting any specific data generation protocol or imaging device that previous works \cite{kalantari2017deep,wei2018deep,chen2018learning} implicitly rely on, 
  hence leading to notably improved real-world generalization.
  \vspace{0.5em}
  \item EnlightenGAN gains remarkable performance by imposing (i) a global-local discriminator structure that handles spatially-varying light conditions in the input image; (ii) the idea of self-regularization, implemented by both the self feature preserving loss and the self-regularized attention mechanism. The self-regularization is critical to our model success, because of the unpaired setting where no strong form of external supervision is available. 
 \vspace{0.5em}
   \item EnlightenGAN is compared with several state-of-the-art methods via comprehensive experiments. The results are measured in terms of visual quality, no-referenced image quality assessment, and human subjective survey. 
  All results consistently endorse the superiority of EnlightenGAN. Moreover, in contrast to existing paired-trained enhancement approaches, EnlightenGAN proves particularly easy and flexible to be adapted to enhancing real-world low-light images from different domains.
\end{itemize}

 

\section{Related Works}

\noindent \textbf{Paired Datasets: Status Quo.} There exist several options to collect a paired dataset of low/normal-light images, but unfortunately none is efficient nor easily scalable. One may fix a camera and then reduce the exposure time in normal-light condition \cite{wei2018deep} or increase exposure time in low-light condition \cite{chen2018learning}. The LOL dataset \cite{wei2018deep}
is so far the only dataset of low/normal-light image pairs taken from real scenes by changing exposure time and ISO. Due to the tedious experimental setup, e.g. the camera needs to be fixed and the object cannot move, etc., it consists of only 500 pairs. Moreover, it may still deviate from the true mapping between natural low/normal-light images. Especially under spatially varying lights, simply increasing/decreasing exposure time may lead to local over-/under-exposure artifacts. 

\begin{figure*}
\centering
\includegraphics[width=17.5cm]{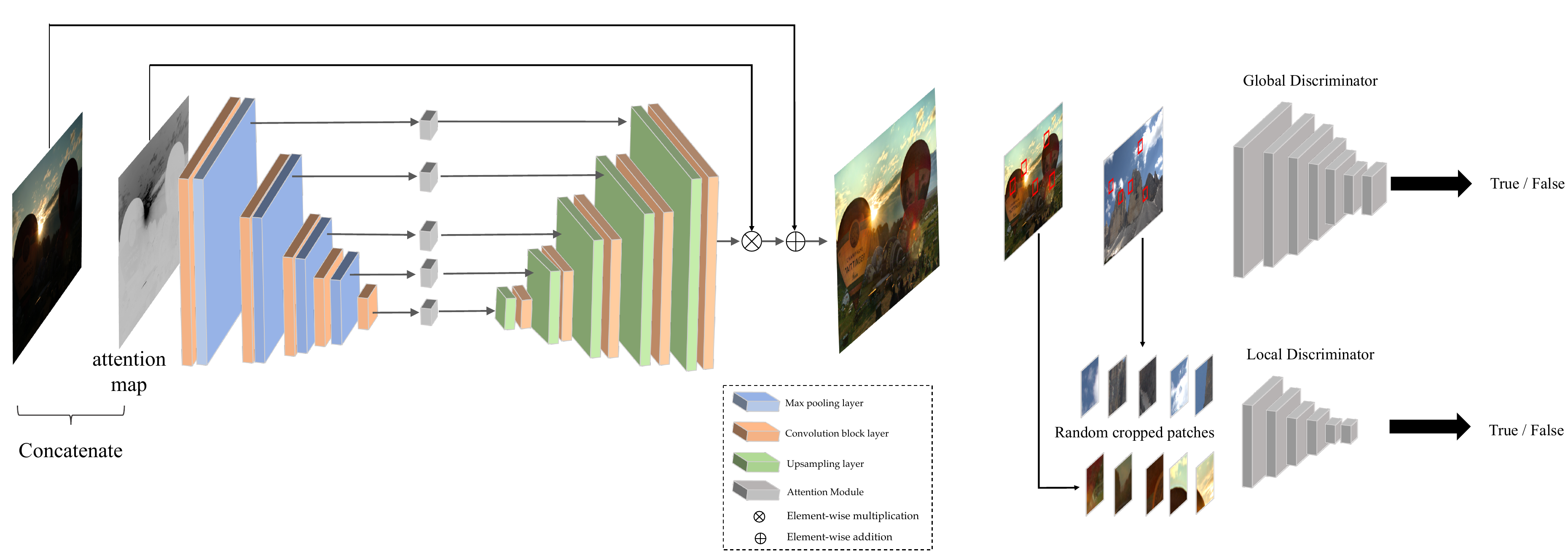}
\caption{The overall architecture of EnlightenGAN. In the generator, each convolutional block consists of two $3$ $\times$ $3$ convolutional layers followed by batch normalization and LeakyRelu. Each attention module has the feature map multiply with a (resized) attention map.}
\label{fig:architecture}
\end{figure*}

In the high-dynamic-ranging (HDR) field, a few works first capture several images at different imperfect light conditions, then align and fuse them into one high-quality image \cite{kalantari2017deep,wu2018deep}. However, they are not designed for the purpose of post-processing only one single low-light image. 

\noindent\textbf{Traditional Approaches.} Low-light image image enhancement has been actively studied as an image processing problem for long, with a few classical methods such as 
the adaptive histogram equalization (AHE) \cite{pizer1987adaptive}, Retinex \cite{land1977retinex} and multi-scale Retinex model \cite{jobson1997multiscale}. 
More recently, \cite{wang2013naturalness} proposed an enhancement algorithm for non-uniform illumination images, utilizing a bi-log transformation to make a balance between details and naturalness. Based on the previous investigation of the logarithmic transformation, Fu \textit{et al.} proposed a weighted variational model \cite{fu2016weighted} to estimate both the reflectance and the illumination from an observed image with imposed regularization terms. 
In \cite{guo2017lime}, a simple yet effective low-light image enhancement (LIME) was proposed, where the illumination of each pixel was first estimated by finding the maximum value in its RGB channels, then the illumination map was constructed by imposing a structure prior. \cite{ren2018joint} introduced a joint low-light image enhancement and denoising model via decomposition in a successive image sequence. \cite{li2018structure} further proposed a robust Retinex model, which additionally considered
a noise map compared with the conventional Retinex
model, to improve the performance of enhancing low-light images
accompanied by intensive noise.

\noindent\textbf{Deep Learning Approaches. }
Existing deep learning solutions mostly rely on paired training, where most low-light images are synthesized from normal images. \cite{lore2017llnet} proposed a stacked auto-encoder (LL-Net) to learn joint denoising and low-light enhancement on the patch level. Retinex-Net in \cite{wei2018deep} provided an end-to-end framework to combine the Retinex theory and deep networks. HDR-Net \cite{gharbi2017deep} incorporated deep networks with the ideas of bilateral grid processing and local affine color transforms with pairwise supervision. A few multi-frame low-light enhancement methods were developed in the HDR domain, such as \cite{kalantari2017deep,wu2018deep,cai2018learning}. 

Lately, \cite{chen2018learning} proposed a ``learning to see in the dark'' model that achieves impressive visual results. However, this method operates directly on
raw sensor data, in addition to the requirement of paired low/normal-light training images. Besides, it focuses more on avoiding the amplified artifacts during low-light enhancement by learning the pipeline of color transformations, demosaicing and denoising, which differs from EnlightenGAN in terms of settings and goal. 

\begin{figure*}[!t]
\centering
\includegraphics[width=17.5cm]{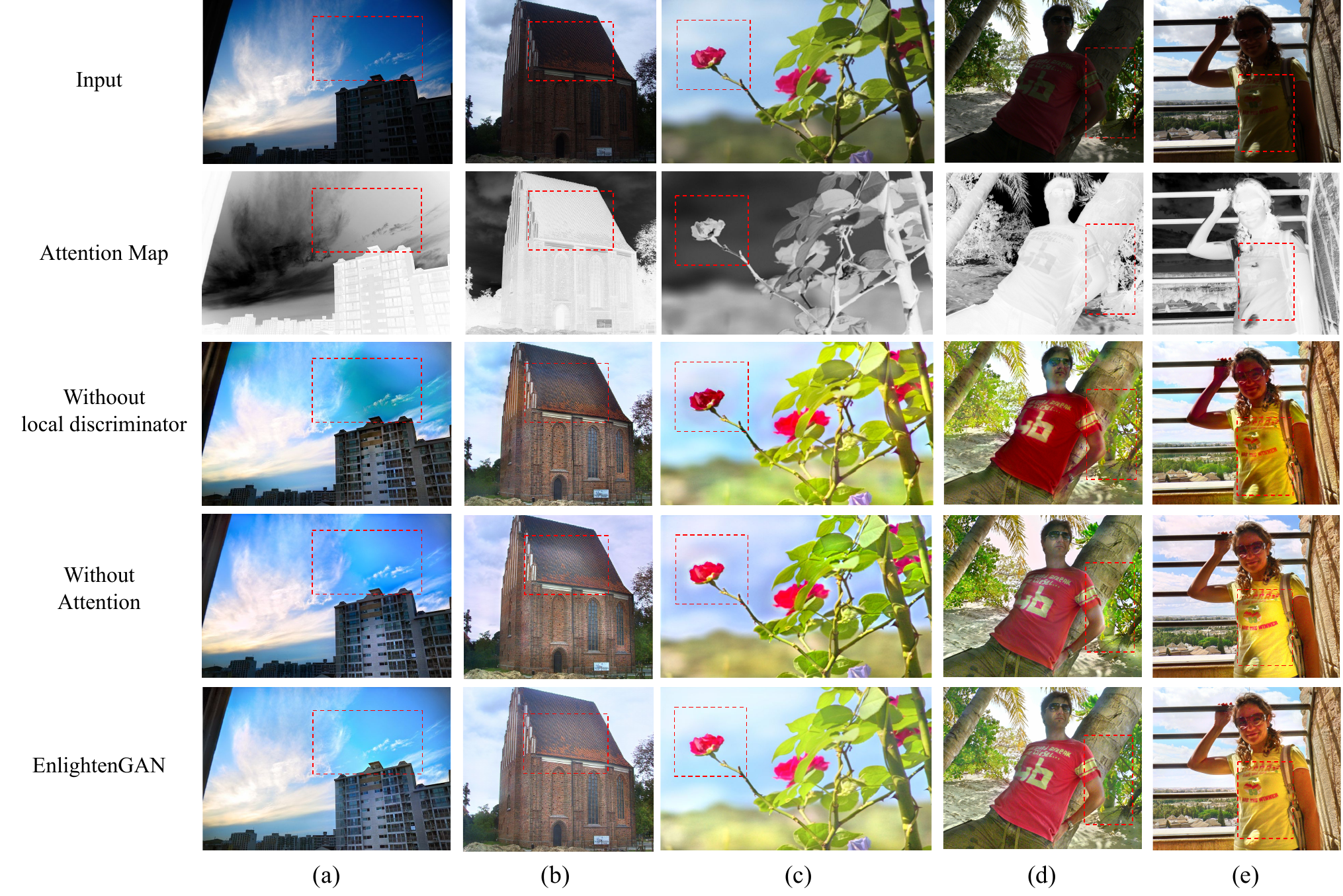}
\caption{Visual comparison from the ablation study of EnlightenGAN. Row 1$\sim$5 display the low-light image inputs, the attention map of input, results from EnlightenGAN with only global discriminator, results from EnlightenGAN without self-regularized attention mechanism, and results from the final version of EnlightenGAN, respectively. Images in Row 3 and 4 suffer from severe color distortion or inconsistency, which are highlighted by bounding boxes. The final version of EnlightenGAN is able to mitigate the above issues and gains the most visually pleasing results. Please zoom in to see the details.}
\label{fig:ablation}
\end{figure*}







\noindent \textbf{Adversarial Learning.}  
GANs \cite{goodfellow2014generative,gong2019autogan} have proven successful in image synthesis and translation. 
When applying GANs to image restoration and enhancement, most existing works use paired training data as well, such as super resolution \cite{ledig2017photo}, artistic style transfer and image editing \cite{yang2019controllable,yang2020deep}, deraining \cite{qian2018attentive} and dehazing \cite{li2018single}. Several unsupervised GANs are proposed to learn inter-domain mappings using adversarial learning and are adopted for many other tasks. \cite{zhu2017unpaired,liu2017unsupervised} adopted a two-way GAN to translate between two different domains by using a cycle-consistent loss with unpaired data
A handful of latest works followed their methodology and applied unpaired training with cycle-consistency to several low-level vision tasks, e.g. dehazing, deraining, super-resolution and mobile photo enhancement  \cite{yang2018towards,yuan122018unsupervised,chen2018deep,jin2018unsupervised}. Different from them, EnlightenGAN refers to unpaired training but with a lightweight one-path GAN structure (i.e., without cycle-consistency), which is stable and easy to train.


\section{Method}
\label{method}
As shown in Fig.~\ref{fig:architecture}, our proposed method adopts an attention-guided U-Net as the generator and uses the dual-discriminator to direct the global and local information. We also use a self feature preserving loss to guide the training process and maintain the textures and structures. In this section we first introduce two important building blocks, i.e., the global-local discriminators and the self feature preserving loss, then the whole network in details. The detailed network architectures are in the supplementary materials.

\subsection{Global-Local Discriminators}
We adopt the adversarial loss to minimize the distance between the real and output normal light distributions. However, we observe that an image-level vanilla discriminator often fails on spatially-varying light images; if the input image has some local area that needs to be enhanced differently from other parts, e.g., a small bright region in an overall dark background, the global image discriminator alone is often unable to provide the desired adaptivity. 

Inspired by previous work \cite{yu2018generative}, to enhance local regions adaptively in addition to improving the light globally, we propose a novel global-local discriminator structure, both using PatchGAN for real/fake discrimination. In addition to the image-level global discriminator, we add a local discriminator by taking randomly cropped local patches from both output and real normal-light images, and learning to distinguish whether they are \textit{real} (from real images) or \textit{fake} (from enhanced outputs). Such a global-local structure ensures all local patches of an enhanced images look like realistic normal-light ones, which proves to be critical in avoiding local over- or under-exposures as our experiments will reveal later.


Furthermore, for the global discriminator, 
we utilize the recently proposed relativistic discriminator structure \cite{jolicoeur2018relativistic} which estimates the probability that real data is more realistic than fake data and also directs the generator to synthesize a fake image that is more realistic than real images. The standard function of relativistic discriminator is:
\begin{equation}
    D_{Ra}(x_r, x_f) = \sigma(C(x_r) - \mathbb{E}_{x_f\sim\mathbb{P}_{\text{fake}}}[C(x_f)]),
\end{equation}
\begin{equation}
    D_{Ra}(x_f, x_r) = \sigma(C(x_f) - \mathbb{E}_{x_r\sim\mathbb{P}_{\text{real}}}[C(x_r)]),
\end{equation}
where $C$ denotes the network of discriminator, $x_r$ and $x_f$ are sampled from the real and fake distribution, $\sigma$ represents the sigmoid function. We slight modify the relativistic discriminator to replace the sigmoid function with the least-square GAN (LSGAN) \cite{mao2017least} loss. 
Finally, the loss functions for the global discriminator $D$ and the generator $G$ are:
\begin{multline}
     \mathcal{L}_D^{Global} = \mathbb{E}_{x_r\sim\mathbb{P}_{\text{real}}}[(D_{Ra}(x_r,x_f) - 1)^2] \\ 
     + \mathbb{E}_{x_f\sim\mathbb{P}_{\text{fake}}}[D_{Ra}(x_f,x_r)^2],
\end{multline}
\begin{multline}
     \mathcal{L}_G^{Global} = \mathbb{E}_{x_f\sim\mathbb{P}_{\text{fake}}}[(D_{Ra}(x_f,x_r) - 1)^2] \\ 
     + \mathbb{E}_{x_r\sim\mathbb{P}_{\text{real}}}[D_{Ra}(x_r,x_f)^2],
\end{multline}
For the local discriminator, we randomly crop 5 patches from the output and real images each time. Here we adopt the original LSGAN as the adversarial loss, as follows:
\begin{multline}
     \mathcal{L}_D^{Local} = \mathbb{E}_{x_r\sim\mathbb{P}_{\text{real-patches}}}[(D(x_r) - 1)^2] \\ 
     + \mathbb{E}_{x_f\sim\mathbb{P}_{\text{fake-patches}}}[(D(x_f) - 0)^2],
\end{multline}
\begin{align}
\mathcal{L}_G^{Local} = \mathbb{E}_{x_r\sim\mathbb{P}_{\text{fake-patches}}}[(D(x_f) - 1)^2],
\end{align}

\begin{figure*}[!ht]
\centering
\includegraphics[width=17.5cm]{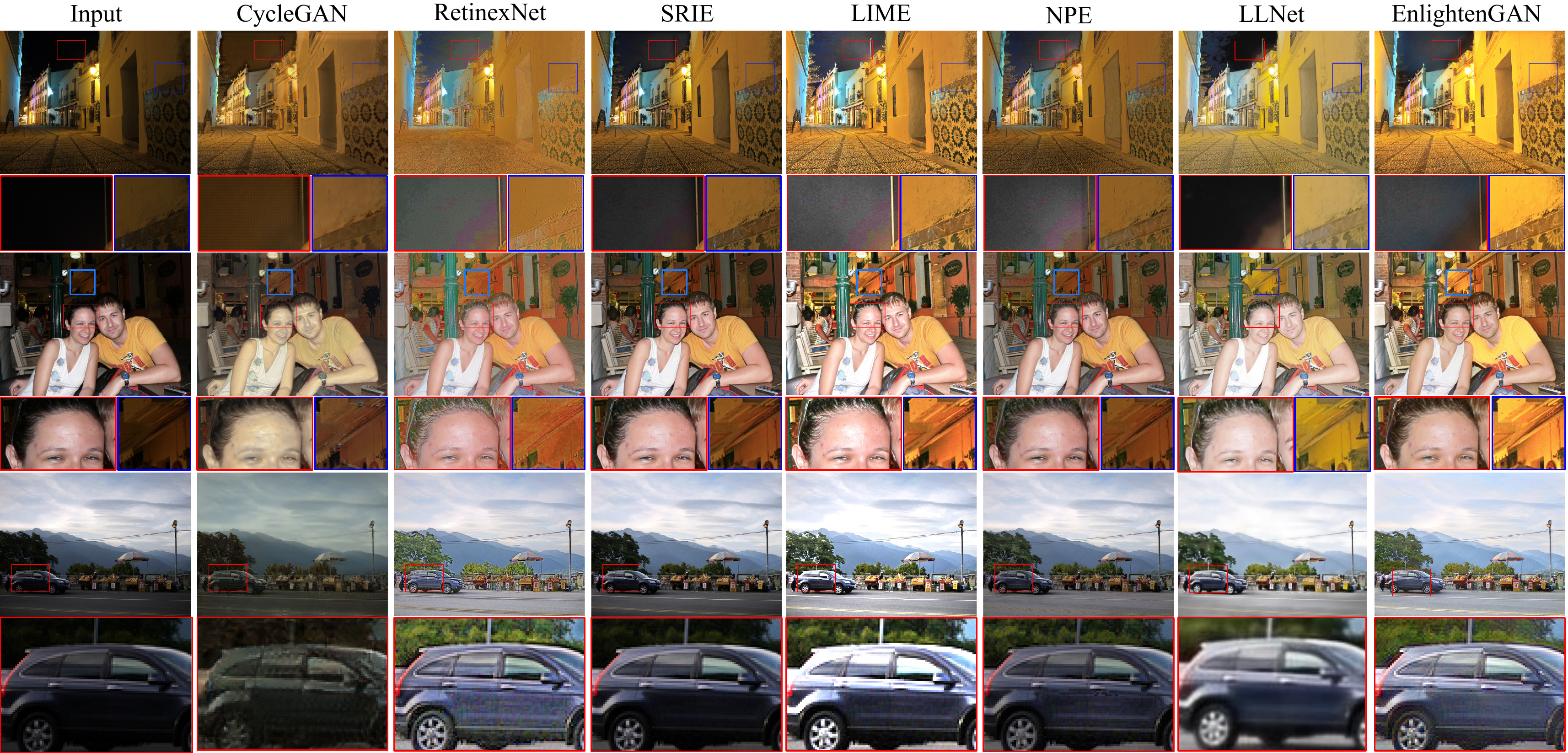}
\caption{Comparison with other state-of-the-art methods. Zoom-in regions are used to illustrate the visual differences. Three examples are listed from the top to the bottom rows. First example: EnlightenGAN successfully suppresses the noise in black sky and produces the best visible details of yellow wall. Second example: NPE and SRIE fail to enhance the background details. LIME introduces over-exposure on the woman's face. LLNet generate severe color distortion. However, EnlightenGAN not only restores the background details but also avoids over-exposure artifacts, distinctly outperforming other methods. Third example: EnlightenGAN produces a visually pleasing result while avoiding over-exposure artifacts in the car and cloud. Others either do not enhance dark details enough or generate over-exposure artifacts. Please zoom in to see more details.}
\label{fig:comparison}
\end{figure*}

\subsection{Self Feature Preserving Loss}
To constrain the perceptual similarity, Johnson \textit{et al.} \cite{johnson2016perceptual} proposed \textit{perceptual loss} by adopting a pre-trained VGG to model feature space distance between images, which was widely adopted to many low-level vision tasks \cite{ledig2017photo,kupyn2017deblurgan}. The common practice constrains the extracted feature distance between the output image and its ground truth. 

In our unpaired setting, we propose to instead constrain the VGG-feature distance between the input low-light and its enhanced normal-light output. This is based on our empirical observation that the classification results by VGG models are not very sensitive when we manipulate the input pixel intensity range, which is concurred by another recent study \cite{richardwebster2018psyphy}. We call it \textit{self feature preserving loss} to stress its \underline{self-regularization} utility to preserve the image content features to itself, before and after the enhancement. That is distinct from the typical usage of the perceptual loss in (paired) image restoration, and is motivated from our unpaired setting too.
Concretely, the self feature preserving loss $L_{SFP}$ is defined as:
\begin{multline}
    \mathcal{L}_{SFP}(I^{L}) = \frac{1}{W_{i,j}H_{i,j}}\sum_{x=1}^{W_{i,j}}\sum_{y=1}^{H_{i,j}}(\phi_{i,j}(I^{L}) 
    - \phi_{i,j}(G(I^{L})))^2,
\end{multline}
where $I^{L}$ denotes the input low-light image and $G(I^{L})$ denotes the generator's enhanced output. $\phi_{i,j}$ denotes the feature map extracted from a VGG-16 model pre-trained on ImageNet. $i$ represents its $i$-th max pooling, and $j$ represents its $j$-th convolutional layer after $i$-th max pooling layer. $W_{i,j}$ and $H_{i,j}$ are the dimensions of the extracted feature maps. By default we choose $i$ = 5, $j$ = 1.

For our local discriminator, the cropped local patches from input and output images are also regularized by a similarly defined self feature preserving loss, $L_{SFP}^{Local}$. 
Furthermore, We add an instance normalization layer \cite{ulyanov2017improved} after the VGG feature maps before feeding into $L_{SFP}$ and $L_{SFP}^{Local}$ in order to stabilize training. The overall loss function for training EnlightenGAN is thus written as:
\begin{equation}
Loss = \mathcal{L}_{SFP}^{Global} + \mathcal{L}_{SFP}^{Local} + \mathcal{L}_G^{Global} + \mathcal{L}_G^{Local},
\end{equation}

\subsection{U-Net Generator Guided with Self-Regularized Attention}
U-Net \cite{ronneberger2015u} has achieved huge success on semantic segmentation, image restoration and enhancement \cite{liu2018image}. By extracting multi-level features from different depth layers, U-Net preserves rich texture information and synthesizes high quality images using multi-scale context information. We adopt U-Net as our generator backbone.

We further propose an easy-to-use attention mechanism for the U-Net generator. Intuitively, in a low-light image of spatially varying light condition, we always want to enhance the dark regions more than bright regions, so that the output image has neither over- nor under-exposure. 
We take the illumination channel $I$ of the input RGB image, normalize it to [0,1], and then use $1-I$ (element-wise difference) as our self-regularized attention map. 
We then resize the attention map to fit each feature map and multiply it with all intermediate feature maps as well as the output image. We emphasize that our attention map is also a form of \underline{self-regularization}, rather than learned with supervision. Despite its simplicity, the attention guidance shows to improve the visual quality consistently. 

Our attention-guided U-Net generator is implemented with 8 convolutional blocks. Each block consists of 
two $3 \times 3$ convolutional layers, followed by LeakyReLu and a batch normalization layer \cite{ioffe2015batch}. At the upsampling stage, we replace the standard deconvolutional layer with one bilinear upsampling layer plus one convolutional layer, to mitigate the checkerboard artifacts. The final architecture of EnlightenGAN is illustrated in the left of Fig. \ref{fig:architecture}. The detailed configuration could be found in the supplementary materials. 

\begin{figure*}[!ht]
\centering
\includegraphics[width=17.5cm]{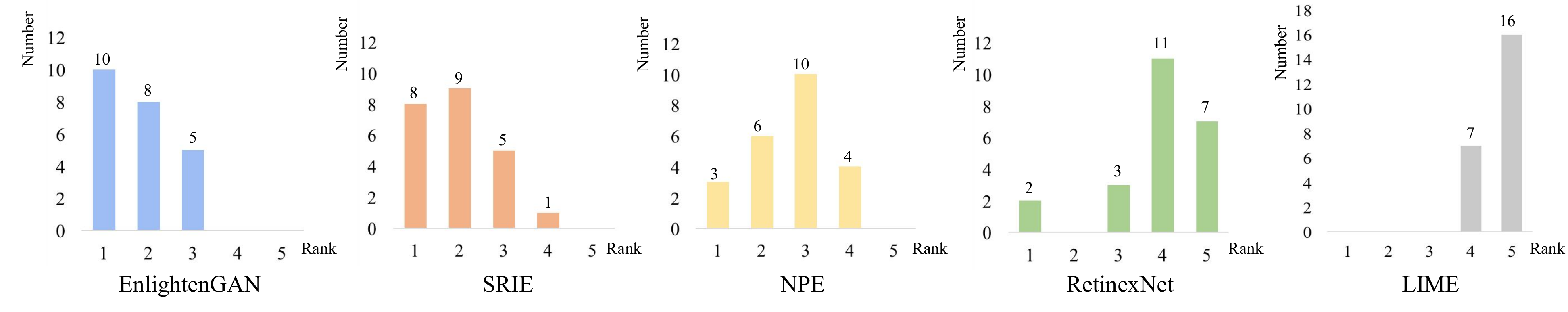}
\caption{The result of five methods in the human subjective evaluation. In each histogram, x-axis denotes the ranking index ($1$ $\sim$ $5$, $1$ represents the highest), and y-axis denotes the number of images in each ranking index. EnlightenGAN produces the most top-ranking images and gains the best performance with the smallest average ranking value.}
\label{fig:table}
\end{figure*}


\section{Experiments}

\subsection{Dataset and Implementation Details}
\label{unpaired}
Because EnlightenGAN has the unique ability to be trained with unpaired low/normal light images, we are enabled to collect a larger-scale unpaired training set, that covers diverse image qualities and contents. We assemble a mixture of 914 low light and 1016 normal light images from several datasets released in \cite{dang2015raise,wei2018deep} and also HDR sources \cite{kalantari2017deep,cai2018learning}, without the need to keep any pair.\footnote{The LOL dataset by \cite{wei2018deep} was a small paired dataset, but we did not use them as pairs for training. An exception is that, we hold out a subset of 50 low/normal light image pairs from LOL \cite{wei2018deep}, as the validation set.} 
Manual inspection and selection are performed to remove images of medium brightness. All these photos are converted to PNG format and resized to $600 \times 400$ pixels. For testing images, we choose those standard ones used in previous works (NPE \cite{wang2013naturalness}, LIME \cite{guo2017lime}, MEF \cite{ma2015perceptual}, DICM \cite{lee2012contrast}, VV, \footnote{https://sites.google.com/site/vonikakis/datasets} etc.).

EnlightenGAN is first trained from the scratch for 100 epochs with the learning rate of 1e-4, followed by another 100 epochs with the learning rate linearly decayed to 0. We use the Adam optimizer and the batch size is set to be 32. Thanks to the lightweight design of one-path GAN without using cycle-consistency, the training time is much shorter than cycle based methods. The whole training process takes 3 hours on 3 Nvidia 1080Ti GPUs.

\subsection{Ablation Study} 

To demonstrate the effectiveness of each component proposed in Sec.~\ref{method}, we conduct several ablation experiments. 
%
Specifically, we design two experiments by removing the components of local discriminator and attention mechanism, respectively. 
As shown in Fig.~\ref{fig:ablation}, the first row shows the input images. The second row shows the attention map of the input images, we can easily observe that the attention map gives a good guideline to the algorithm by which region should be enhanced more while others should be enhanced less. The third row shows the image produced by EnlightenGAN with only global discriminator to distinguish between low-light and normal-light images. The fourth row is the result produced by EnlightenGAN which does not adopt self-regularized attention mechanism and uses U-Net as the generator instead. The last row is produced by our proposed version of EnlightenGAN.


The enhanced results in the third row and the fourth row tend to contain local regions of severe color distortion or under-exposure, namely, the sky over the building in Fig.\ref{fig:ablation}(a), the roof region in  Fig.\ref{fig:ablation}(b), the left blossom in Fig.\ref{fig:ablation}(c), the boundary of tree and bush in Fig.\ref{fig:ablation}(d), and the T-shirt in Fig.\ref{fig:ablation}(e). 
In contrast, the results of the full EnlightenGAN contain realistic color and thus more visually pleasing, which validates the effectiveness of the global-local discriminator design and self-regularized attention mechanism. More images are in the supplementary materials.

\subsection{Comparison with State-of-the-Arts}

In this section we compare the performance of EnlightenGAN with current state-of-the-art methods. We conduct a list of experiments including visual quality comparison, human subjective review and no-referenced image quality assessment (IQA), which are elaborated on next. 

\subsubsection{Visual Quality Comparison}
\label{visual}
We first compare the visual quality of EnlightenGAN with several recent competing methods. Results are demonstrated in Fig.~\ref{fig:comparison}, where the first column shows the original low-light images, and the second to fifth columns are the images enhanced by: a vanilla CycleGAN 
\cite{zhu2017unpaired} trained using our unpaired training set, RetinexNet \cite{wei2018deep}, SRIE \cite{fu2016weighted}, LIME  \cite{guo2017lime}, NPE \cite{wang2013naturalness}, LLNet \cite{lore2017llnet}, and CycleGAN \cite{zhu2017unpaired}. The last column shows the results produced by EnlightenGAN.

We next zoom in on some details in the bounding boxes. 
LIME easily leads to over-exposure artifacts, which makes the results distorted and glaring with the some information missing.  
The results of SRIE and NPE are generally darker compared with others. 
CycleGAN and RetinexNet generate unsatisfactory visual results in terms of both brightness and naturalness. 
In contrast, EnlightenGAN successfully not only learns to enhance the dark area but also preserves the texture details and avoids over-exposure artifacts.
More results are shown in the supplementary materials.

\begin{figure*}[!ht]
\centering   
\includegraphics[width=18cm]{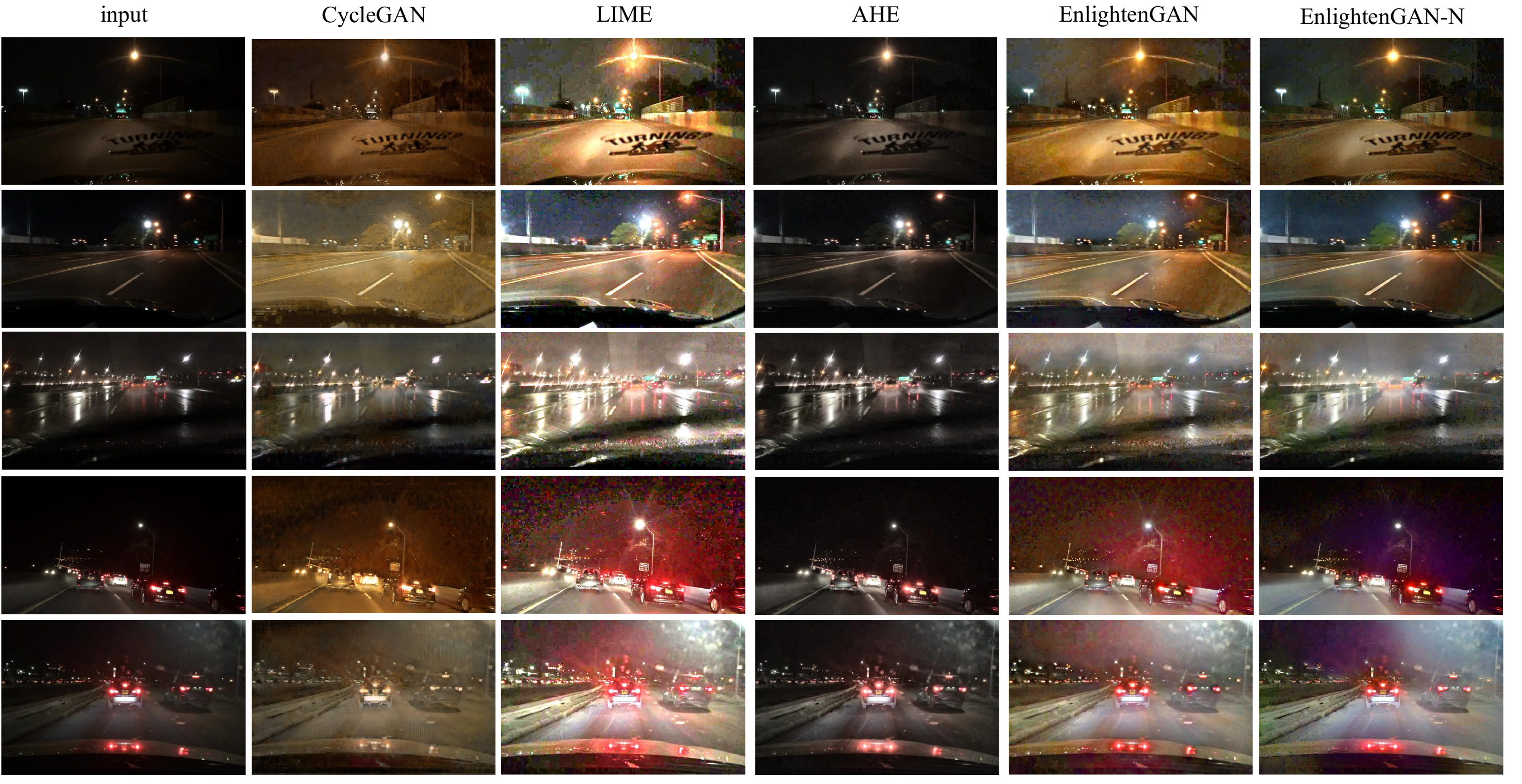}
\caption{Visual comparison of the results on the BBD-100k dataset \cite{yu2018bdd100k}. EnlightenGAN-N is the domain-adapted version of EnlightenGAN, which generates the most visually pleasing results with noise suppressed. Please zoom in to see the details.}
\vspace{-1em}
\label{fig:BBD}
\end{figure*}

\subsubsection{No-Referenced Image Quality Assessment}

\begin{table}[!thb]
\begin{center}
\caption{NIQE scores on the whole testing set (All) and each subset (MEF, LIME, NPE, VV, DICM) respectively. Smaller NIQE indicates more perceptually favored quality.}
\label{table:niqe}
\small
\setlength{\tabcolsep}{1.3mm}{
\begin{tabular}{|c|c|c|c|c|c|c|}
\hline
Image set & MEF & LIME & NPE & VV & DICM & All  \\
\hline\hline
Input & 4.265 & 4.438 & 4.319 & 3.525 & 4.255 & 4.134\\
\hline
LLNet & 4.845 & 4.940 & 4.78 & 4.446 & 4.809 & 4.751\\
\hline
CycleGAN & 3.782 & 3.276 & 4.036 & 3.343 & 3.560 & 3.554\\
\hline
RetinexNet & 4.149 & 4.420 & 4.485 & 2.602 & 4.200 & 3.920\\ 
\hline
LIME & 3.720 & 4.155 & 4.268 & \textbf{2.489} & 3.846 & 3.629\\
\hline
SRIE & 3.475 & 3.788 & 3.986 & 2.850 & 3.899 & 3.650\\
\hline
NPE & 3.524 & 3.905 & \textbf{3.953} & 2.524 & 3.760 & 3.525\\
\hline

EnlightenGAN & \textbf{3.232} & \textbf{3.719} & 4.113 & 2.581 & \textbf{3.570} & \textbf{3.385}\\
\hline
\end{tabular}}
\end{center}

\end{table}


We adopt Natural Image Quality Evaluator (NIQE) \cite{mittal2013making}, a well-known no-reference image quality assessment for evaluating real image restoration without ground-truth, to provide quantitative comparisons. The NIQE results on five publicly available image sets used by previous works (MEF, NPE, LIME, VV, and DICM) are reported in Table~\ref{table:niqe}: a lower NIQE value indicates better visual quality. EnlightenGAN wins on three out of five sets, and is the best in terms of overall averaged NIQE. This further endorses the superiority of EnlightenGAN over current state-of-the-art methods in generating high-quality visual results.

\subsubsection{Human Subjective Evaluation}\label{subjective}
We conduct a human subjective study to compare the performance of EnlightenGAN and other methods. We randomly select 23 images from the testing set. For each image, it is first enhanced by five methods (LIME, RetinexNet, NPE, SRIE, and EnlightenGAN). We then ask 9 subjects to independently compare the five outputs in a pairwise manner. Specifically, each time a human subject is displayed with a pair of images randomly drawn from the five outputs, and is asked to evaluated which one has better quality. The human subjects are instructed to consider the: 1) whether the images contain visible noise; 2) whether the images contain over- or under-exposure artifacts; and 3) whether the images show nonrealistic color or texture distortions. Next, we fit a Bradley-Terry model  \cite{bradley1952rank} to estimate the numerical subjective scores so that the five methods can be ranked, using the exactly same routine as described in previous works \cite{li2019benchmarking}. As a result, each method is assigned with rank 1-5 on that image. We repeat the above for all 23 images.



Fig.~\ref{fig:table} displays the five histograms, each of which depicts the rank distributions that a method receives on the 23 images. For example, EnlightGAN has been ranked the 1st (i.e., the highest subjective score) on 10 out of 23 images, the 2nd for 8 images, and the 3rd for 5 images. By comparing the five histograms, it is clear that EnlightenGAN produces the overall most favored results by human subjects, with an average ranking of 1.78 over 23 images. RetinexNet and LIME are not well scored, because of causing many over-exposures and sometimes amplifying the noise.

\subsection{Adaptation on Real-World Images}
\label{bbd}

Domain adaptation is an indispensable factor for real-world generalizable image enhancement. 
The unpaired training strategy of EnlightenGAN allows us to directly learn to enhance real-world low-light images from various domains, where there is \textbf{no paired normal-light} training data or even \textbf{no normal-light} data from the same domain available. We conduct experiments using low-light images from a real-world driving dataset, Berkeley Deep Driving (BBD-100k) \cite{yu2018bdd100k}, to showcase this \textbf{unique advantage} of EnlightenGAN in practice.


We pick 950 night-time photos (selected by mean pixel intensity values smaller than 45) from the BBD-100k set as the low-light training images, plus 50 low-light images for hold-out testing. Those low-light images suffer from severe artifacts and high ISO noise.
We then compare two EnlightenGAN versions trained on different normal-light image sets, including: 1) the pre-trained EnlightenGAN model as described in Sec.~\ref{unpaired}, without any adaptation for BBD-100k; 
2) \textbf{EnlightenGAN-N}: a domain-adapted version of EnlightenGAN, which uses BBD-100k low-light images from the BBD-100k dataset for training, while the normal-light images are still the high-quality ones from our unpaired dataset in Sec.~\ref{unpaired}. We also include a traditional method, Adaptive histogram equalization (AHE), and a pre-trained LIME model for comparison, and an unsupervised approach CycleGAN.

As shown in Fig.~\ref{fig:BBD}, the results from LIME suffer from severe noise amplification and over-exposure artifacts, while AHE does not enhance the brightness enough. The unsupervised approach CycleGAN generate very low quality due to its unstability. The original EnlightenGAN also leads to noticeable artifacts on this unseen image domain. In comparison, EnlightenGAN-N produces the most visually pleasing results, striking an impressive balance between brightness and artifact/noise suppression. Thanks to the unpaired training, EnlightenGAN could be easily adapted into EnlightenGAN-N without requiring any supervised/paired data in the new domain, which greatly facilitates its real-world generalization.

\subsection{Pre-Processing for Improving Classification}
\label{classification}

Image enhancement as pre-processing for improving subsequent high-level vision tasks has recently received increasing attention \cite{kupyn2017deblurgan,li2017aod,liu2018improved,kupyn2019deblurgan}, with a number of benchmarking efforts  \cite{li2019benchmarking,yuan2019ug,li2019single,yang2020advancing}. We investigate the impact of light enhancement on the \textit{extremely dark} (\textbf{ExDark}) dataset \cite{Exdark}, which was specifically built for the task of low-light image recognition. The classification results after light enhancement could be treated as an indirect measure on semantic information preservation, as \cite{kupyn2017deblurgan,li2019benchmarking} suggested. 

The ExDark dataset consists of 7,363 low-light images, including 3000 images in training set, 1800 images in validation set and 2563 images in testing set, annotated into 12 object classes. We use its testing set only, applying our pretrained EnlightenGAN as a pre-processing step, followed by passing through another ImageNet-pretrained ResNet-50 classifier. Neither domain adaption nor joint training is performed. The high-level task performance serves as a fixed semantic-aware metric for enhancement results.

In the low-light testing set, using EnlightenGAN as pre-processing improves the classification accuracy from 22.02\% (top-1) and 39.46\% (top-5), to 23.94\% (top-1) and 40.92\% (top-5) after enhancement. That supplies a side evidence that EnlightenGAN preserves semantic details, in addition to producing visually pleasing results. We also conduct experiment using LIME and AHE. LIME improves the accuracy to 23.32\% (top-1) and 40.60\% (top-5), while AHE obtains to 23.04\% (top-1) and 40.37\% (top-5). 


\section{Conclusion}
\vspace{-0.5em}
In this paper, we address the low-light enhancement problem with a novel and flexible unsupervised framework. The proposed EnlightenGAN operates and generalizes well without any paired training data. 
The experimental results on various low light datasets show that our approach outperforms multiple state-of-the-art approaches under both subjective and objective metrics. Furthermore, we demonstrate that EnlightenGAN can be easily adapted on real noisy low-light images and yields visually pleasing enhanced images. Our future work will explore how to control and adjust the light enhancement levels based on user inputs in one unified model. Due to the complicacy of light enhancement, we also expect integrate algorithm with sensor innovations.

{\small
\bibliographystyle{unsrt}
\bibliography{egbib}
}

\begin{IEEEbiography}
[{\includegraphics[width=1in,height=1.25in,clip,keepaspectratio]{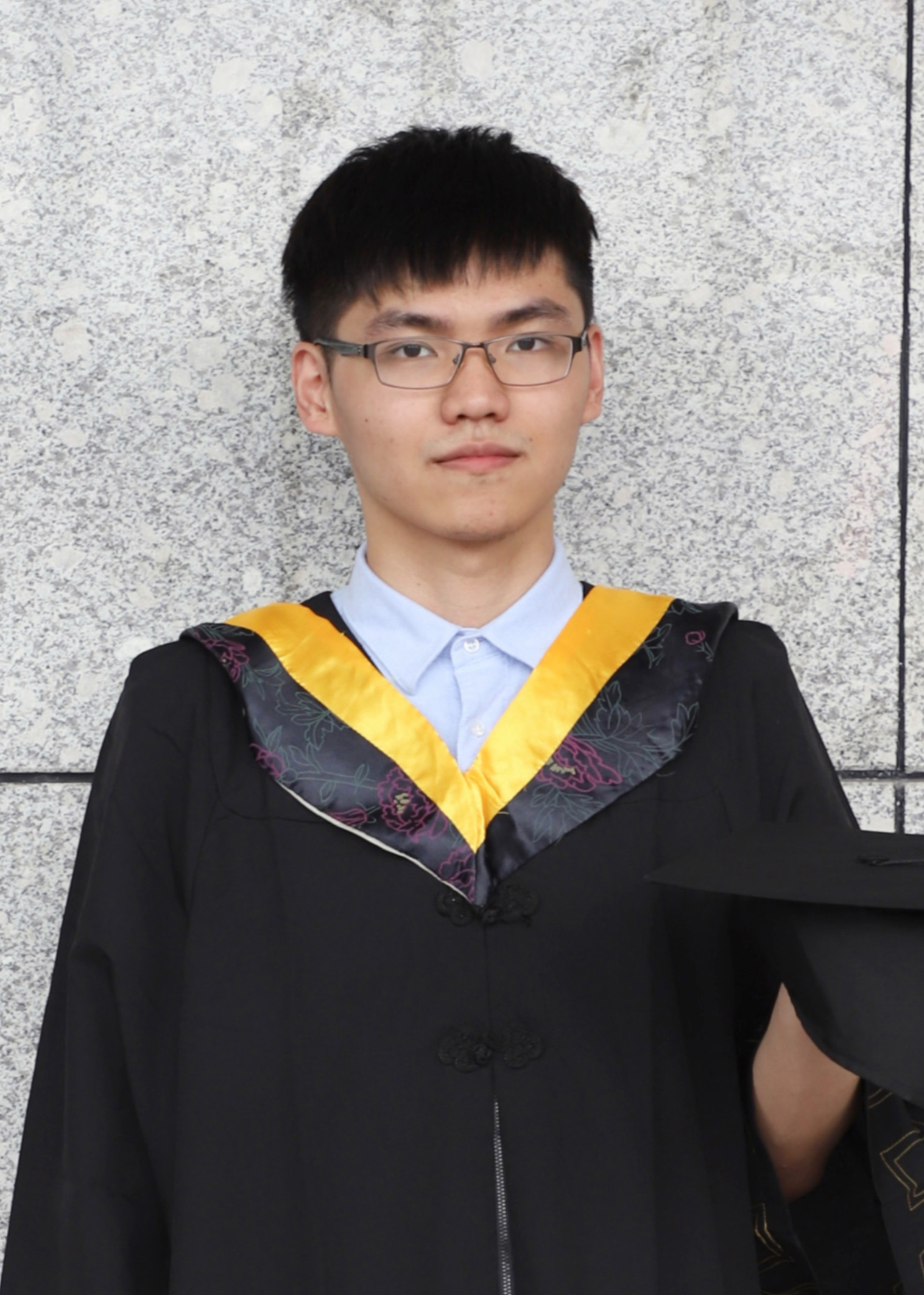}}]{Yifan Jiang}
is currently a Ph.D. student at The University of Texas at Austin. He received his bachelor degree from the Huazhong University of Science and Technology in 2019. His research interests include computer vision and deep learning. Now he mainly works on the area of generative models and image enhancement.
\end{IEEEbiography}

\begin{IEEEbiography}
[{\includegraphics[width=1in,height=1.25in,clip,keepaspectratio]{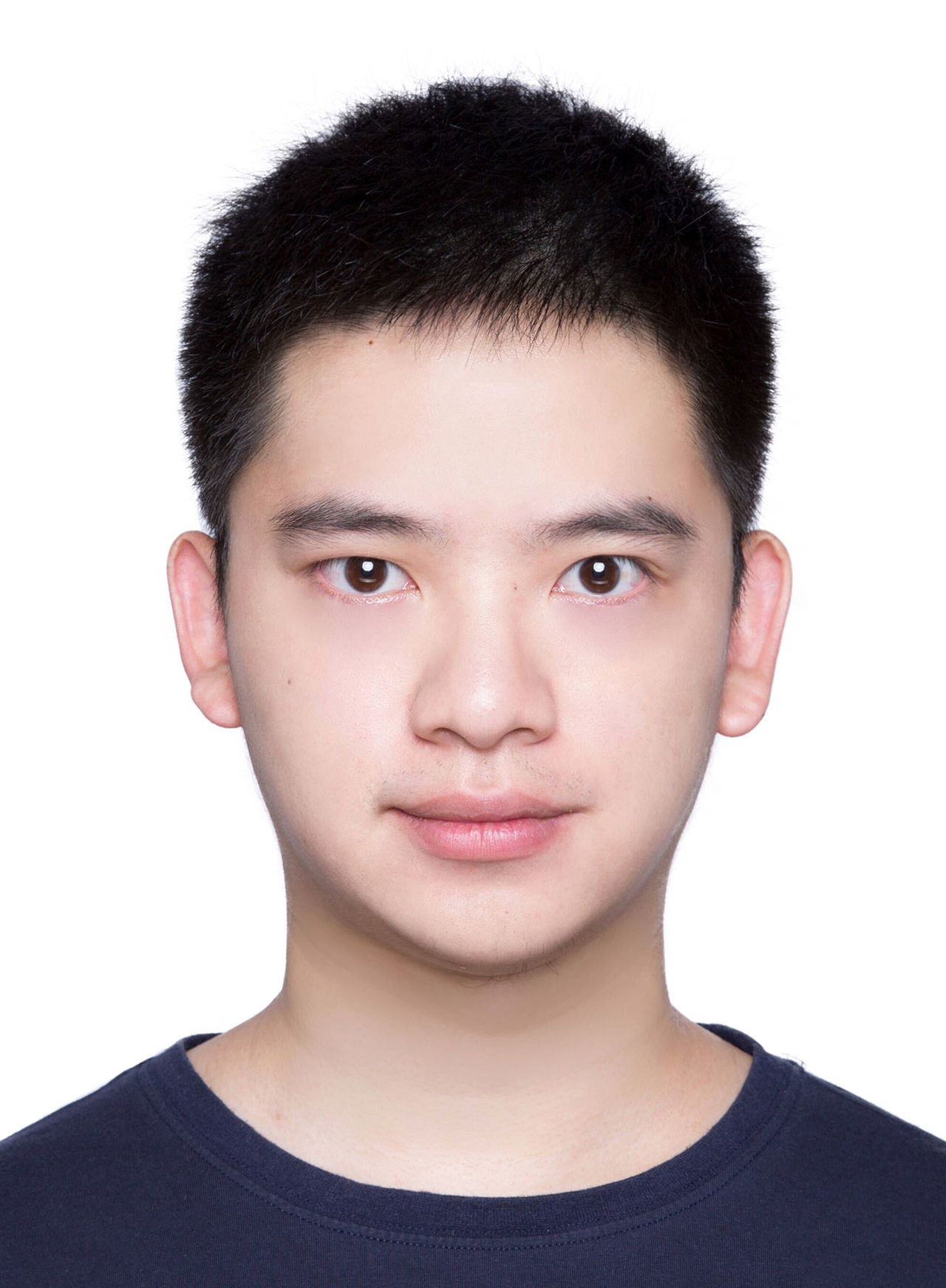}}]{Xinyu Gong}
 is currently a Ph.D. student of Electrical and Computer Engineering at The University of Texas at Austin. He receives his bachelor degree from University of Electronic Science and Technology of China in 2018. His research interests include AutoML and generative models.
\end{IEEEbiography}

\begin{IEEEbiography}
[{\includegraphics[width=1in,height=1.25in,clip,keepaspectratio]{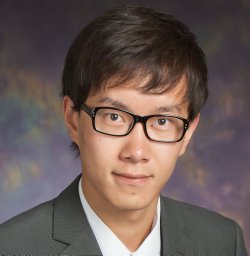}}]{Ding Liu}
 received the Ph.D. degree from the University of Illinois at Urbana–
Champaign, USA, in 2018. He is currently a
Research Scientist with Bytedance Inc., Mountain View, CA, USA. His research experience
encompasses low-level vision problems, including
image/video restoration and enhancement. He has
broad research interests in the area of computer
vision, image processing. and deep learning.
\end{IEEEbiography}

\begin{IEEEbiography}
[{\includegraphics[width=1in,height=1.25in,clip,keepaspectratio]{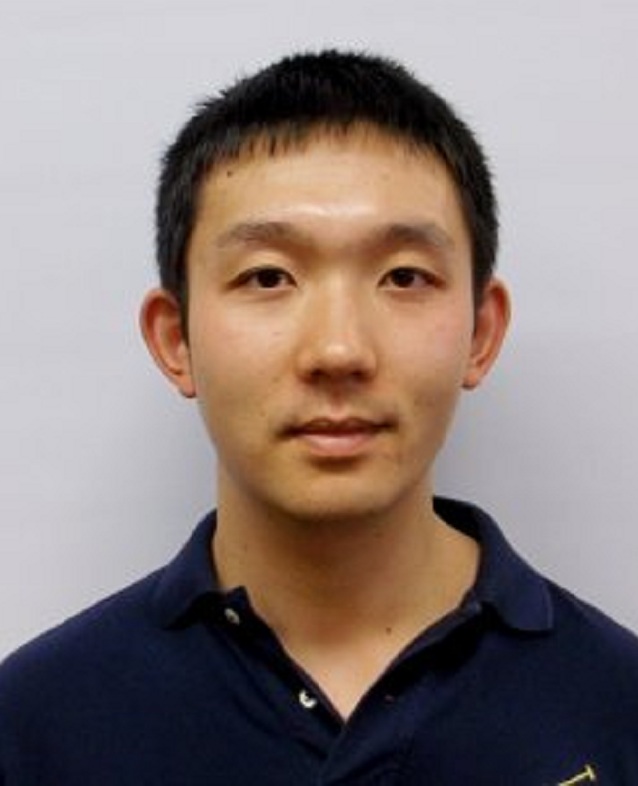}}]{Yu Cheng}
is  a  researcher  at  Microsoft.  Before  that,  he  spent  three  years as  a  research  staff  member  at  IBM  T.J.  Watson  Research  Center.  He  got  his  Ph.D.  from Northwestern University in 2015 and bachelors from Tsinghua University in 2010. His research is about deep learning in general, with specific interests  in  the  deep  generative  model,  model compression, and reinforcement learning. He is also  interested  in  solving  real-world  problems of  computer  vision  and  natural  language  processing. He regularly serves on the program committees of top-tier AI conferences such as NIPS, ICML, ICLR, CVPR and ACL.
\end{IEEEbiography}

\begin{IEEEbiography}
[{\includegraphics[width=1in,height=1.25in,clip,keepaspectratio]{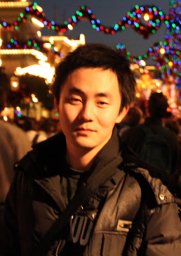}}]{Chen Fang}
is  a  researcher  at  Bytedance Inc.  Before  that,  he was a research scientist at Adobe Research. He  got  his  Ph.D.  from Dartmouth in 2015 and bachelors from Hunan University in 2010. His research interests include image understanding, image search, image editing, generative models
\end{IEEEbiography}

\begin{IEEEbiography}
[{\includegraphics[width=1in,height=1.25in,clip,keepaspectratio]{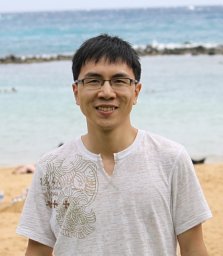}}]{Xiaohui Shen}
 is a Researcher at ByteDance
AI Lab. Before that, he was a Senior Research
Scientist at Adobe Research. He obtained his
PhD degree from the Department of EECS at
Northwestern University, and received the MS
and BS degrees from the Department of Automation at Tsinghua University, China. His research
interests include computer vision and deep learning.
\end{IEEEbiography}

\begin{IEEEbiography}
[{\includegraphics[width=1in,height=1.25in,clip,keepaspectratio]{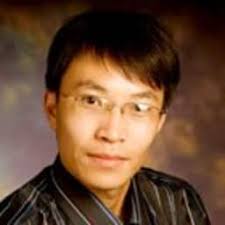}}]{Jianchao Yang} received the M.S. and Ph.D.
degrees from the ECE Department, University of
Illinois at Urbana–Champaign, under the supervision
of Prof. Thomas Huang. He was a Research Scientist with
Adobe Research. He is currently a Director with Bytedance Inc. He has authored over 80
technical papers over a wide variety of topics on top
tier conferences and journals, with Google scholar
citation over 12 000 times. His research focuses
on computer vision, deep learning, and image and
video processing. He received the Best Student Paper
award from ICCV 2010, the Classification Task Prize in PASCAL VOC 2009,
first position for object localization using external data for ILSVRC ImageNet
2014, and third place in the WebVision Challenge 2017. He serves as the
Workshop Chair of the ACM MM 2017.
\end{IEEEbiography}

\begin{IEEEbiography}
[{\includegraphics[width=1in,height=1.25in,clip,keepaspectratio]{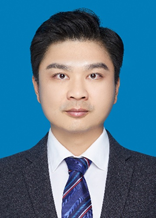}}]{Pan Zhou}
is currently a full professor with Hubei Engineering Research Center on Big Data Security, School of Cyber Science and Engineering, Huazhong University of Science and Technology. He received his Ph.D. in the School of Electrical and Computer Engineering at the Georgia Institute of Technology (Georgia Tech) in 2011, Atlanta, USA. He was a senior technical member at Oracle Inc., from 2011 to 2013. His current research interests include security and privacy, big data analytics, machine learning, and information networks.
\end{IEEEbiography}

\begin{IEEEbiography}
[{\includegraphics[width=1in,height=1.25in,clip,keepaspectratio]{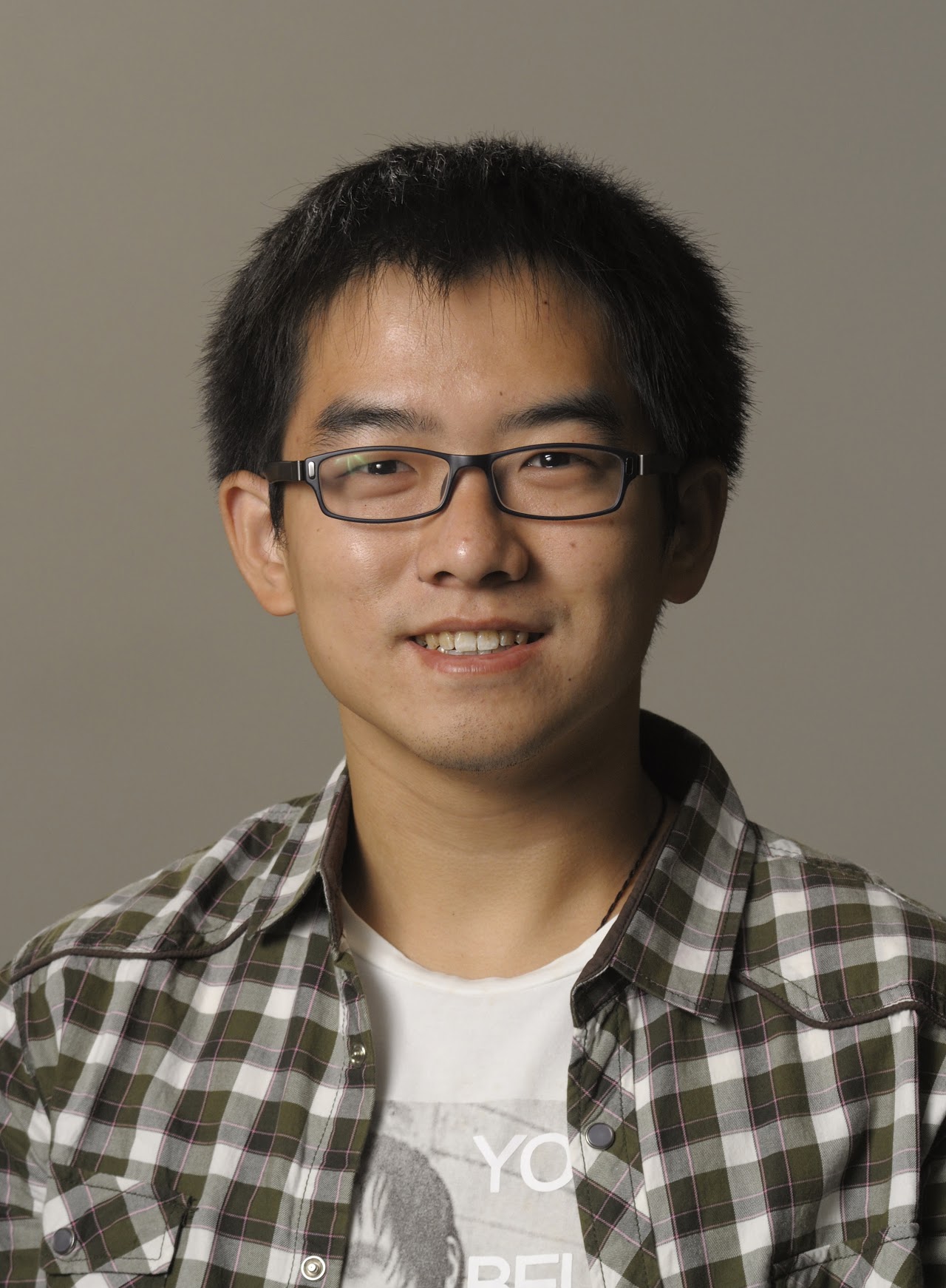}}]{Zhangyang Wang}
 is currently an Assistant Professor at The University of Texas at Austin. He was an Assistant
Professor  at Texas A\&M University, from 2017 to 2020. He received his Ph.D. degree the ECE Department, University of
Illinois at Urbana–Champaign, under the supervision
of Prof. Thomas Huang. Prof. Wang is broadly interested in the fields of machine learning, computer vision, optimization, and their interdisciplinary applications. His latest interests focus on automated machine learning (AutoML), learning-based optimization, machine learning robustness, and efficient deep learning.
\end{IEEEbiography}



%








\end{document}